\documentclass{article}
\pdfoutput=1
\usepackage{spconf,amsmath,graphicx,bbm,booktabs}
\usepackage[numbers,sort]{natbib}
\usepackage[hidelinks]{hyperref}
\usepackage{tabularx}
\newcolumntype{L}[1]{>{\raggedright\arraybackslash}m{#1}}
\newcolumntype{C}[1]{>{\centering\arraybackslash}p{#1}}

\title{ACTIVEMATCH: END-TO-END SEMI-SUPERVISED ACTIVE REPRESENTATION LEARNING}

\name{Xinkai Yuan$^*$, Zilinghan Li$^*$, Gaoang Wang$^+$
\thanks{$^*$These two authors have equal contributions to this work.}
\thanks{$^+$Corresponding author.}}
\address{Zhejiang University-University of Illinois at Urbana-Champaign Institute, Zhejiang University, China}
\begin{document}
%
\maketitle
\begin{abstract}
Semi-supervised learning (SSL) is an efficient framework that can train models with both labeled and unlabeled data, but may generate ambiguous and non-distinguishable representations when lacking adequate labeled samples.
With human-in-the-loop, active learning can iteratively select informative unlabeled samples for labeling and training to improve the performance in the SSL framework. However, most existing active learning approaches rely on pre-trained features, which is not suitable for end-to-end learning.
To deal with the drawbacks of SSL, in this paper, we propose a novel end-to-end representation learning method, namely ActiveMatch, which combines SSL with contrastive learning and active learning to fully leverage the limited labels. Starting from a small amount of labeled data with unsupervised contrastive learning as a warm-up, ActiveMatch then combines SSL and supervised contrastive learning, and actively selects the most representative samples for labeling during the training, resulting in better representations towards the classification. Compared with MixMatch and FixMatch with the same amount of labeled data, we show that ActiveMatch achieves the state-of-the-art performance, with 89.24\% accuracy on CIFAR-10 with 100 collected labels, and 92.20\% accuracy with 200 collected labels. 
\end{abstract}
%
\section{Introduction}
\label{sec:intro}
Deep neural networks have achieved great performance in many computer vision applications, such as image classification \cite{SupBaseline}, object detection \cite{NN_APP2}, and instance segmentation \cite{NN_APP4}. It is observed that the great performance of deep neural networks usually requires a large amount of labeled data for training \cite{EMP2}. However, labeling data is expensive and time-consuming, so it is demanding to develop methods that can train the model using only a small set of labeled data.

The semi-supervised learning (SSL) framework alleviates the demand for large amounts of labels.
It utilizes techniques such as pseudo-labeling \cite{PseudoLabel, PseudoLabel2}, consistency regularization \cite{ConsistentReg2,ConsistentReg3,ConsistentReg4}, or the combination of these two to leverage unlabeled data \cite{MixMatch,ReMixMatch,UDA, FixMatch, SelfMatch}. For pseudo-labeling-based approaches, the predictions on the unlabeled data are treated as pseudo labels 
to train the models if the predictions exceed a threshold. Consistency regularization employs unlabeled samples by minimizing distances between the predictions of different augmented copies from the same original sample. Advanced methods like MixMatch \cite{MixMatch}, FixMatch \cite{FixMatch} and SelfMatch \cite{SelfMatch} combine those two techniques and achieve high accuracy with few labels.

Contrastive learning is another prevailing method to leverage unlabeled data and is widely used to learn sample representations for downstream tasks \cite{MatInfo,SimCLR,MoCo,SimCLR2,CSI}. Unsupervised contrastive learning tries to make distances between representations of different augmented copies of the same sample smaller and enlarge the distances between representations of different samples. When labels are also available, supervised contrastive learning, which minimizes representation distances for samples from the same class and repels the representations among different classes, is also employed to help to learn a more generative representation \cite{SupervisedCL}.

Active learning aims to select the most representative samples from the unlabeled dataset for oracle to label, and there have been many algorithms for selection according to the uncertainty, diversity, consistency, and density of samples \cite{ActiveSurvey,UncertaintyActive,AL_ImgClass,semi_active,AL_OD}. 
Those selected representative labeled samples help to improve the model performance the most. With only a small amount of actively selected labels, active learning can usually achieve promising performance. 
However, such active learning based approaches usually require reliable pre-trained features for sample selection, and are not suitable for end-to-end representation learning.

In this paper, following the SSL setting, we propose a novel representation learning method, namely ActiveMatch, which combines contrastive learning and active learning into an end-to-end learning framework.
Starting from a few randomly selected labels, ActiveMatch employs unsupervised contrastive learning to initialize the representation. To utilize the information from labeled samples, the network is then trained with supervised contrastive learning and SSL, and actively selects the most uncertain samples measured on the unlabeled set for labeling during the training until the desired amount of labels has been obtained. The framework is built on FixMatch \cite{FixMatch}, a state-of-the-art (SOTA) representation learning method with SSL. Unlike FixMatch, we combine supervised contrastive learning to better represent samples with labeled contrastive pairs. Moreover, the active sample selection can overcome the limitation of training with a small amount of labeled data, and further boost the classification performance. With the same amount of labeled samples, ActiveMatch outperforms previous SSL methods on benchmarks including CIFAR-10, CIFAR-100 \cite{CIFAR}, and SVHN \cite{SVHN}. The contributions of our paper are summarized as follows:
\begin{itemize}
\setlength{\itemsep}{0pt}

\setlength{\parsep}{0pt}

\setlength{\parskip}{0pt}
    \item Proposed ActiveMatch is a novel representation learning approach that combines SSL, contrastive learning, and active learning to address the issue of training with a small amount of labeled data.
    \item Unlike several other methods which use contrastive learning as pre-training and then fine-tune the network based on SSL, ActiveMatch uses an end-to-end training scheme, which simplifies the training process and helps to improve the accuracy.
    \item ActiveMatch outperforms previous SSL methods on standard benchmarks such as CIFAR-10, CIFAR-100, and SVHN with only a few labeled samples.
\end{itemize}

\section{Method}

\begin{figure*}[t]
  \begin{minipage}[t]{1.0\linewidth}
    \centering
    \centerline{\includegraphics[width=15cm]{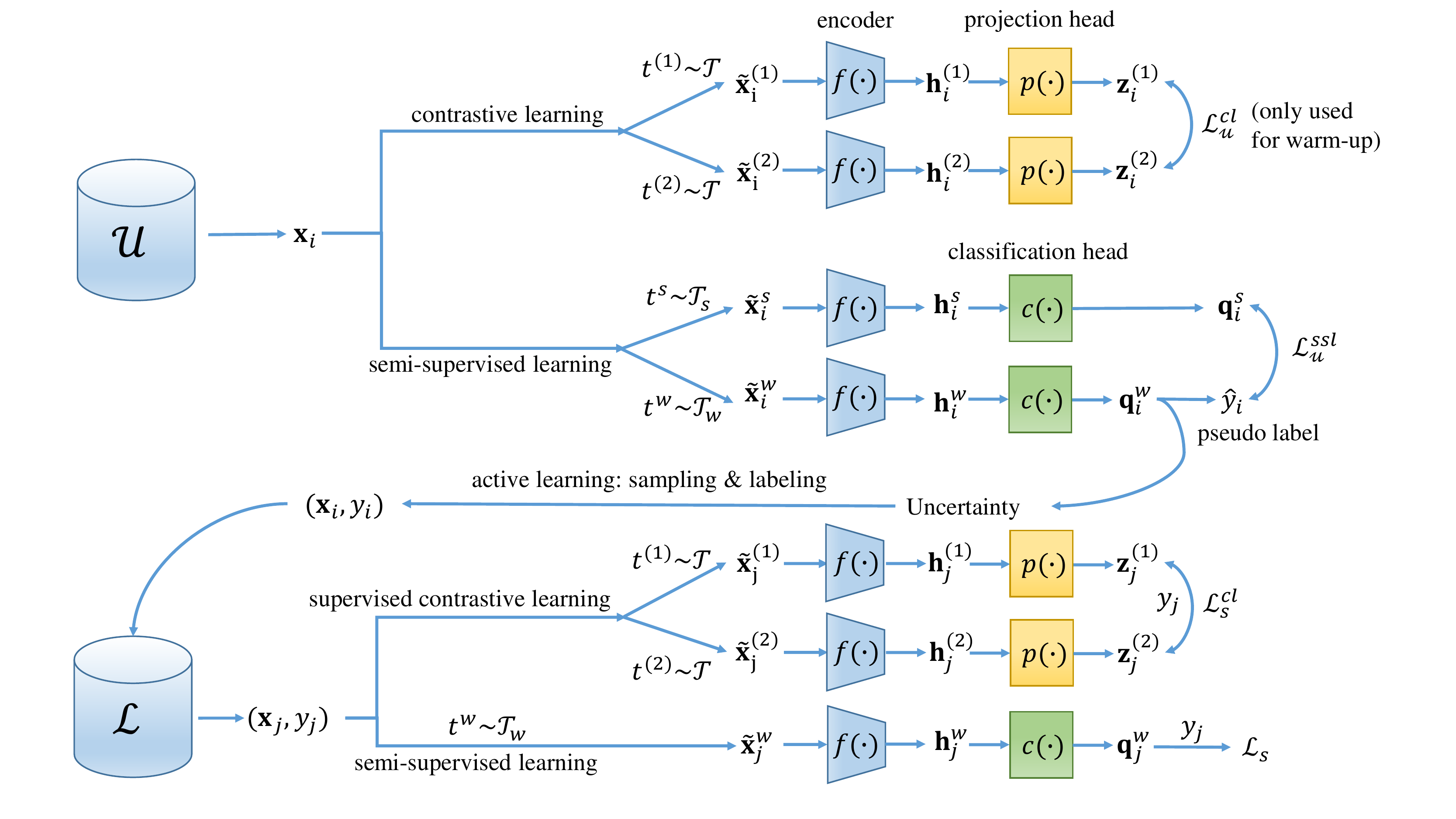}}
  \end{minipage}
\caption{Overview of ActiveMatch. Both labeled and unlabeled data employ contrastive learning and semi-supervised learning to train the network in ActiveMatch. Based on the uncertainty of predictions of weakly augmented images, active learning selects the representative image for labeling and appends the newly labeled image into the labeled set.}
\vspace{-3mm}
\label{fig:overview}
\end{figure*}

The overview of ActiveMatch is shown in Fig. \ref{fig:overview}. The figure indicates that ActiveMatch is composed of two main parts, one unsupervised part which trains the network with a large unlabeled dataset $\mathcal{U}$ and one supervised part which trains with a small labeled dataset $\mathcal{L}$. Active learning connects these two parts by periodically selecting the most representative samples from $\mathcal{U}$ for labeling according to the uncertainty of samples. 
Both supervised and unsupervised parts consist of two learning methods, contrastive learning which aims to optimize the representation, and semi-supervised learning which aims to improve the class prediction accuracy.

\subsection{Unsupervised Contrastive Learning for Unlabeled Samples}
The underlying idea of contrastive learning is to learn an encoder generating representations such that the representations of positive samples have small distances and representations of negative samples have large distances. Given an image $\mathbf{x}$, let $\{\mathbf{x}_+\}$ be the set of positive samples of $\mathbf{x}$, $\{\mathbf{x}_-\}$ be the set of negative samples of $\mathbf{x}$ and $r(\cdot)$ be the function to generate the representation of $\mathbf{x}$. In addition, we employ cosine similarity function $\text{sim}(\mathbf{x}_1,\mathbf{x}_2)={\mathbf{x}_1\cdot \mathbf{x}_2}/{\left\|\mathbf{x}_1\right\|\left\|\mathbf{x}_2\right\|}$ to measure the distance between representations. Then the loss for contrastive learning can be defined in the following way:
\begin{equation}\label{GeneralConLoss}
  \begin{split}
    &\mathcal{L}^{cl}(\mathbf{x}, \{\mathbf{x}_+\}, \{\mathbf{x}_-\}) = \\
    &\frac{-1}{|\{\mathbf{x}_+\}|}\log\frac{\sum_{\mathbf{x}'\in\{\mathbf{x}_+\}}\exp(\text{sim}(r(\mathbf{x}),r(\mathbf{x}'))/\tau)}{\sum_{\mathbf{x}'\in\{\mathbf{x}_+\}\cup\{\mathbf{x}_-\}}\exp(\text{sim}(r(\mathbf{x}),r(\mathbf{x}'))/\tau)},
  \end{split}
\end{equation}
where $\tau$ is the temperature parameter. 

For unsupervised contrastive learning, like SimCLR \cite{SimCLR}, the set $\{\mathbf{x}_+\}$ is obtained
by augmenting the same image in different ways. Let $\mathcal{B}_{\mathcal{U}}=\{\mathbf{x}_i\}_{i=1}^{B_U}$ be a batch of unlabeled images, for each image $\mathbf{x}_i$, we apply two different augmentations $t^{(1)}, t^{(2)}\in\mathcal{T}$ to it and obtain $\tilde{\mathbf{x}}_i^{(1)}=t^{(1)}(\mathbf{x}_i),\ \tilde{\mathbf{x}}_i^{(2)}=t^{(2)}(\mathbf{x}_i)$. Both $t^{(1)}$ and $t^{(2)}$ are the combinations of two simple augmentations such as random crop, random color distortion, and Gaussian blur. Denote augmented unlabeled image set as $\tilde{\mathcal{B}}_{\mathcal{U}}=\{\tilde{\mathbf{x}}_i^{(1)}, \tilde{\mathbf{x}}_i^{(2)}\}_{i=1}^{B_U}$. Then for $\mathbf{x}_i$, its augmentations $\tilde{\mathbf{x}}_i^{(1)}$ and $\tilde{\mathbf{x}}_i^{(2)}$ are considered to be positive pairs, while all others in $\tilde{\mathcal{B}}_{\mathcal{U}}\backslash \{\tilde{\mathbf{x}}_i^{(1)}, \tilde{\mathbf{x}}_i^{(2)}\}$ are considered to be negative samples. In this way, the loss function for unsupervised contrastive learning is defined as:
\begin{equation}\label{UnsupervisedConLoss}
  \begin{split}
    \mathcal{L}_{\mathcal{U}}^{cl}(\mathcal{B}_{\mathcal{U}}) = \frac{1}{2B_U}\mathop{\sum}\limits_{i=1}^{B_U}&\mathcal{L}^{cl}(\tilde{\mathbf{x}}_i^{(1)},\{\tilde{\mathbf{x}}_i^{(2)}\}, \tilde{\mathcal{B}}_{\mathcal{U}}\backslash \{\tilde{\mathbf{x}}_i^{(1)}, \tilde{\mathbf{x}}_i^{(2)}\})\\
    +&\mathcal{L}^{cl}(\tilde{\mathbf{x}}_i^{(2)},\{\tilde{\mathbf{x}}_i^{(1)}\}, \tilde{\mathcal{B}}_{\mathcal{U}}\backslash \{\tilde{\mathbf{x}}_i^{(1)}, \tilde{\mathbf{x}}_i^{(2)}\}).
  \end{split}
\end{equation}

\subsection{Supervised Learning for Labeled Samples}


First, we apply cross-entropy as the supervised loss on labeled samples. Specifically, for batch $\mathcal{B}_{\mathcal{L}}=\{(\mathbf{x}_j,y_j)\}_{j=1}^{B_L}$, the supervised loss is defined as:
\begin{equation}\label{SupervisedSSLLoss}
  \mathcal{L}_{\mathcal{S}}(\mathcal{B}_{\mathcal{L}})=\frac{1}{B_L}\mathop{\sum}\limits_{j=1}^{B_L}H(y_j, \mathbf{q}_j),
\end{equation}
where $\mathbf{q}_j$ is the prediction and $H(\cdot)$ represents the cross-entropy.

To enlarge the discrimination between different classes, we also employ the supervised contrastive loss to learn the relationships among samples. 
Like unsupervised contrastive learning, each image $\mathbf{x}_j$ gets augmented twice as well. However, since labels are available, the augmentations of images belonging to the same class are considered to be positive. Let $\mathcal{B}_{\mathcal{L}}=\{(\mathbf{x}_j, y_j)\}_{j=1}^{B_L}$ be the labeled batch and $\tilde{\mathcal{B}}_{\mathcal{L}}=\{\tilde{\mathbf{x}}_j^{(1)},\tilde{\mathbf{x}}_j^{(2)}\}_{j=1}^{B_L}$ be its augmentations. Then for image $\mathbf{x}_j$, the positive set is defined as $\mathcal{S}_j = \{\tilde{\mathbf{x}}_k^{(1)}, \tilde{\mathbf{x}}_k^{(2)}|y_k=y_j\}$, and all other images in $\tilde{\mathcal{B}}_{\mathcal{L}}$ are considered to be negative. In this way, the loss function for supervised contrastive loss is defined as:
\begin{equation}\label{SupervisedConLoss}
  \begin{split}
    \mathcal{L}_{\mathcal{S}}^{cl}(\mathcal{B}_{\mathcal{L}}) = \frac{1}{2B_L}\mathop{\sum}\limits_{j=1}^{B_L}&\mathcal{L}^{cl}(\tilde{\mathbf{x}}_j^{(1)},\mathcal{S}_j\backslash\{\tilde{\mathbf{x}}_j^{(1)}\}, \tilde{\mathcal{B}}_{\mathcal{L}}\backslash\mathcal{S}_j)\\
    +&\mathcal{L}^{cl}(\tilde{\mathbf{x}}_j^{(2)},\mathcal{S}_j\backslash\{\tilde{\mathbf{x}}_j^{(2)}\}, \tilde{\mathcal{B}}_{\mathcal{L}}\backslash\mathcal{S}_j).
  \end{split}
\end{equation}

\subsection{Semi-Supervised Learning with Pseudo Labels}
Semi-supervised learning (SSL) leverages the information from labeled set to unlabeled set in representation learning. In addition, the classification head is also trained with SSL. Following FixMatch \cite{FixMatch}, the details of SSL are demonstrated as follows. 

For an unlabeled image $\mathbf{x}_i$, it is augmented by one weak augmentation and one strong augmentation to obtain $\tilde{\mathbf{x}}_i^w,\ \tilde{\mathbf{x}}_i^s$ respectively. Specifically, the weak augmentation randomly flips the image horizontally with a probability of $50\%$, and the strong augmentation is the combination of RandAugment \cite{RandAugment} and Cutout \cite{Cutout}. The network then generates predictions $\mathbf{q}_i^w,\ \mathbf{q}_i^s$ for $\tilde{\mathbf{x}}_i^w,\ \tilde{\mathbf{x}}_i^s$. If the maximum value of $\mathbf{q}_i^w$ exceeds a confidence threshold $c$, then $\hat{y}_i=\text{argmax}(\mathbf{q}_i^w)$ is considered as the pseudo label for $\mathbf{x}_i$ and will be used to compute the cross-entropy loss for $\mathbf{q}_i^s$. Given an unlabeled batch $\mathcal{B}_{\mathcal{U}}=\{\mathbf{x}_i\}_{i=1}^{B_U}$, the loss function of SSL for unlabeled images is defined as:
\begin{equation}\label{UnsupervisedSSLLoss}
  \mathcal{L}_{\mathcal{U}}^{ssl}(\mathcal{B}_{\mathcal{U}})=\frac{1}{B_U}\mathop{\sum}\limits_{i=1}^{B_U}\mathbbm{1}(\text{max}(\mathbf{q}_i^w)>c)H(\hat{y}_i, \mathbf{q}_i^s),
\end{equation}
where $\mathbbm{1}(\cdot)$ is the indicator function which evaluates to 1 if and only if the condition inside is true.

\subsection{Model Training with Active Learning}

We employ an uncertainty-based approach, margin sampling, which is efficient and has low computation complexity. Typically, margin sampling selects the sample from the unlabeled set with the smallest probability difference between the most probable and second probable classes. Active learning connects supervised learning and unsupervised learning in ActiveMatch. Combined with contrastive learning loss and SSL loss, the total loss is expressed as follows,
\begin{equation}
  \mathcal{L} = \lambda_1\mathcal{L}_{\mathcal{U}}^{cl}+\lambda_2\mathcal{L}_{\mathcal{S}}+\lambda_3\mathcal{L}_{\mathcal{S}}^{cl}+\lambda_4\mathcal{L}_{\mathcal{U}}^{ssl},
\end{equation}
where $\lambda$s are the weights of different loss terms.

\section{Experiments}
\begin{table*}[t]
\small
\centering
\caption{Accuracy comparison on CIFAR-10, CIFAR-100, and SVHN.}\label{tab:dataset}
\begin{tabular}{llccclccclcc}
\toprule
        &  & \multicolumn{3}{c}{CIFAR-10} &  & \multicolumn{3}{c}{CIFAR-100} &  & \multicolumn{2}{c}{SVHN} \\ \cmidrule{1-1} \cmidrule{3-5} \cmidrule{7-9} \cmidrule{11-12} 
Method  &  & 50 labels       & 100 labels    & 200 labels    &  & 500 labels      & 1000 labels     & 2000 labels   &  & 100 labels     & 200 labels       \\ \cmidrule{1-1} \cmidrule{3-5} \cmidrule{7-9} \cmidrule{11-12} 
Supervised \cite{SupBaseline} &  & \textbf{96.23}    & \textbf{96.23}   & \textbf{96.23}   &  & \textbf{80.27}    & \textbf{80.27}    & \textbf{80.27}   &  & \textbf{98.41}  & \textbf{98.41}    \\ \midrule
MixMatch \cite{MixMatch} &  &   57.36       &    71.82     &    86.76     &  &      18.76    &    32.64      &    48.08     &  &     80.33   &   95.62            \\ 
FixMatch \cite{FixMatch} &  &    75.16      &    88.99     &    91.39     &  &      36.62    &    47.64      &    56.12     &  &     \textbf{97.00}   &     \textbf{97.55}          \\
\textbf{ActiveMatch} &  &     \textbf{78.25}     &    \textbf{89.24}     &    \textbf{92.20}     &  &   \textbf{40.26}       &     \textbf{52.20}     &     \textbf{60.33}    &  &    96.97    &     97.47          \\ \bottomrule 
\end{tabular}
\vspace{-5mm}
\end{table*}
\subsection{Experimental Setup}
We evaluate the performance of ActiveMatch on SSL benchmarks such as CIFAR-10, CIFAR-100 \cite{CIFAR}, and SVHN \cite{SVHN}, and compare the performance with supervised learning and two other SSL methods with different amounts of labels. To make a fair comparison, 
other SSL methods randomly select the same amount of labeled samples for training.

ResNet-18 \cite{NN_APP3} is used as the encoder backbone. For the learning rate, we employ cosine learning rate decay \cite{CosLR} whose learning rate is defined as $lr={lr}_0\cdot\cos(7\pi k/16K)$, where ${lr}_0$ is the initial learning rate, $K$ is the total number of training steps and $k$ is the current training step. For Active Learning, ActiveMatch starts with a small labeled set $\mathcal{L}$ with $n_0$ labels and warms up the network using only unsupervised contrastive learning loss $\mathcal{L}_{\mathcal{U}}^{cl}$ for $t_{wp}$ epochs.
Active learning starts to sample images for every $B_{smp}$ batches until reaching the desired number of labels. The reasons for the warm-up are twofold. First, the warm-up makes the image representations from the encoder meaningful to ensure the representativeness of the selected sample from active learning. Second, it also improves the accuracy of pseudo labels used in SSL.
The hyperparameters we use in experiments are $\lambda_3=0.08,\ \lambda_2=\lambda_4=1,\ \tau=0.07,\ B_L=64,\ B_U=7\cdot64=448,\ c = 0.95,\ lr_0 = 0.03,\ t_{wp}=15$. The initial number of labels $n_0$ and the number of batches between each sampling $B_{smp}$ depend on the dataset and the total number of labels to be collected, so their values will be specified in the following subsections. It should be mentioned that all accuracy shown in the following tables are the average values of three independent experiments.

\subsection{Results for CIFAR-10, CIFAR-100, and SVHN}
Table \ref{tab:dataset} shows the accuracy comparison of different methods on CIFAR-10, CIFAR-100, and SVHN. For CIFAR-10, $n_0=10$, $B_{smp}=128$ for 50 labels, $n_0=20$, $B_{smp}=64$ for 100 labels, and $n_0=40$, $B_{smp}=32$ for 200 labels. Additionally, for the case with 50 labels sampled in total, $t_{wp}=5$. For CIFAR-100, $n_0=100$ for 500 labels, $n_0=200$ for 1000 labels, and $n_0=400$ for 2000 labels. $B_{smp}=4$ for all three cases. Results indicate that ActiveMatch achieves the state-of-the-art performance on CIFAR-10 and CIFAR-100. 
Since SVHN is a relatively simple dataset, ActiveMatch achieves almost the same performance with FixMatch, and also very close to the fully supervised performance.

\subsection{Qualitative Results}

\begin{figure}[htb]

\begin{minipage}[b]{1.0\linewidth}
  \centering
  \centerline{\includegraphics[width=9.5cm]{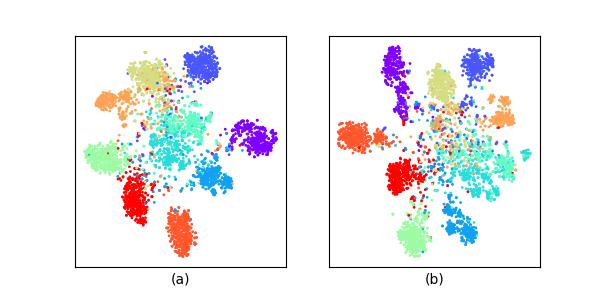}}
\end{minipage}
\caption{t-SNE plots of representations learned by two SSL methods on the first 10 classes of CIFAR-100. (a): ActiveMatch trained with 2000 labels. (b): FixMatch trained with 2000 labels.}
\label{fig:tsne}
\vspace{-5mm}
\end{figure}

To compare the representations learned by ActiveMatch and FixMatch and visualize them, we apply t-SNE \cite{van2008visualizing} to reduce the dimensionality of the representations coming from the encoder and plot them in Fig. \ref{fig:tsne}. We selected images from the first ten classes on the CIFAR-100 and randomly selected 400 samples from each class. The results show that the representations obtained by our network work better on clustering than FixMatch, particularly for samples around the center of each sub-figure.

\subsection{Ablation Study}
In ActiveMatch, we propose that the use of active learning helps to train the SSL model more efficiently by selecting representative images to label. In Table \ref{tab:ablation}, we compare the classification accuracy of our network with and without active learning. The results indicate that with the use of active learning, the accuracy can be improved by approximately $1\%\sim2\%$. ActiveMatch also employs supervised contrastive learning, and in Table \ref{tab:ablation}, we also show that supervised contrastive learning helps to improve the model prediction accuracy by around $2\%\sim3\%$.
\begin{table}[!htbp]
  \centering
  \caption{Effect of active learning (AL) on ActiveMatch model (CIFAR-10).}\label{tab:ablation}	\begin{tabular}{lccc}
		\toprule  
		Method & 100 labels & 200 labels\\
		\midrule 
    ActiveMatch  & \textbf{89.24} & \textbf{92.20}\\
    ActiveMatch (without AL) & 87.76 & 89.91\\
    ActiveMatch (without $\mathcal{L}_{\mathcal{S}}^{cl}$) & 87.34 & 88.89\\
		\bottomrule 
	\end{tabular}
    \vspace{-5mm}
\end{table}

\section{Conclusion}
In this paper, we propose ActiveMatch, which is an end-to-end semi-supervised learning (SSL) method combining SSL, contrastive learning, and active learning for learning representations. ActiveMatch leverages a relatively small labeled dataset and a large unlabeled dataset to achieve good performance on the image classification task. Experiments show that ActiveMatch achieves the state-of-the-art on SSL benchmarks CIFAR-10, CIFAR-100, and SVHN. Additionally, ActiveMatch shows how active learning can help to improve the performance of SSL. We believe that it is worth further investigations on how advanced active learning algorithms can provide more benefits to semi-supervised learning.

\bibliographystyle{IEEEbib}
\small
\setlength{\bibsep}{0pt}
\bibliography{refs}

\end{document}